\documentclass[letterpaper, 10 pt, conference]{ieeeconf}  

\IEEEoverridecommandlockouts                              

\usepackage{booktabs}
\usepackage{gensymb}
\usepackage{graphicx,amsmath,amssymb,cite}
\usepackage{xcolor}
\usepackage{breakurl} 
\usepackage{caption}
\usepackage{subcaption}
\usepackage{balance}
\usepackage{algorithm}
\usepackage{algpseudocode}
\usepackage{tikz}

\newcommand\copyrighttext{%
\footnotesize
\textcopyright~2026 IEEE. Personal use of this material is permitted.
Permission from IEEE must be obtained for all other uses, in any current or future media, including reprinting/republishing this material for advertising or promotional purposes, creating new collective works, for resale or redistribution to servers or lists, or reuse of any copyrighted component of this work in other works.
}
\newcommand\copyrightnotice{%
\begin{tikzpicture}[remember picture,overlay]
\node[anchor=south,yshift=10pt] at (current page.south)
{\fbox{\parbox{\dimexpr 0.75\textwidth-\fboxsep-\fboxrule\relax}{\copyrighttext}}};
\end{tikzpicture}%
}

\newcommand{\wfourdia}{0.205}

\title{\LARGE \bf
Force-Based Offset Estimation for Keyed Peg-in-Hole Assembly Using Local Gaussian Process Regression
}

\author{%
Chandra Yuvesh Aubeeluck$^{1}$,
Abilash Philip Madavath$^{1}$,
Augustin Raju$^{1}$\\
Nicolas Pyschny$^{1}$,
Felix Hackelöer$^{1}$,
Florian Zwanzig$^{1}$ 
\thanks{*This work is supported by the InnoFaktur project through the European Regional Development Fund (EFRE
20500002).}
\thanks{$^{1}$Institute of Mechanical Engineering, Cologne University of Applied Sciences, Campus Gummersbach, Germany
        {\tt\small chandra\_yuvesh.aubeeluck@th-koeln.de}}
}

\begin{document}
\maketitle
\copyrightnotice

\begin{abstract}
Key--keyway assembly tasks impose strict geometric constraints and are highly sensitive to grasp pose deviations in uncertain environments.
This work presents a force-based offset estimation method for keyed peg-in-hole assembly, embedded within a perception--validation--insertion pipeline.
Residual misalignment is estimated directly from wrist force/torque measurements using a local KNN--Gaussian Process hybrid regressor.
The framework distinguishes between two contact regimes, hard collision and guided chamfer insertion, and routes inference to a dedicated model for each. Regime classification is achieved via a contact-window duration threshold.
KNN combined with a deterministic search using the results of a post-grasp monocular visual validation contributes to an increased accuracy of the regressor model.
This approach achieves accurate radial offset estimation in chamfered peg insertion, during a keypoint detection-based pick and place application.
Experiments using the integrated force/torque sensor of a collaborative robot arm showed an increase in insertion success rate from 67\% to 87\% after the pipeline was applied.
\end{abstract}
\textit{\textbf{keywords: machine learning, robot dynamics and control, complex assembly, precision placement}}

\section{Introduction and Related Work} \label{sec:introduction}
\cite{tang2016autonomous}
Precision placement in robotics is classically represented by the peg-in-hole problem, where a component must be inserted into a geometrically corresponding counterpart under tight tolerances.
Keyed hole setups, such as motor shafts and torque-driven components, extend this problem by imposing additional circumferential orientation constraints, making validated insertion without collision considerably more demanding~\cite{whitney1982force}.
When geometric variability is present, systems typically rely on visual pose estimation and force feedback.
AI-models for pose-estimation provide reliable pre-grasp localization ~\cite{kostas2025reviewing}, while two-stage pipelines using external cameras refine object pose via robot kinematics and multi-view keypoint consistency~\cite{yuan2023single,duverney2026affordable}.
During insertion, interaction between the peg and hole generates characteristic force/torque (F/T) signatures that encode contact-state and residual misalignment information \cite{rusautocon2021}.
This work investigates continuous radial and angular offset estimation from wrist F/T signatures and the integration of post-grasp pose validation into trajectory planning, enabling deterministic insertion correction rather than heuristic search.

\begin{figure}[t]
    \centering
    \includegraphics[width=0.75\columnwidth]{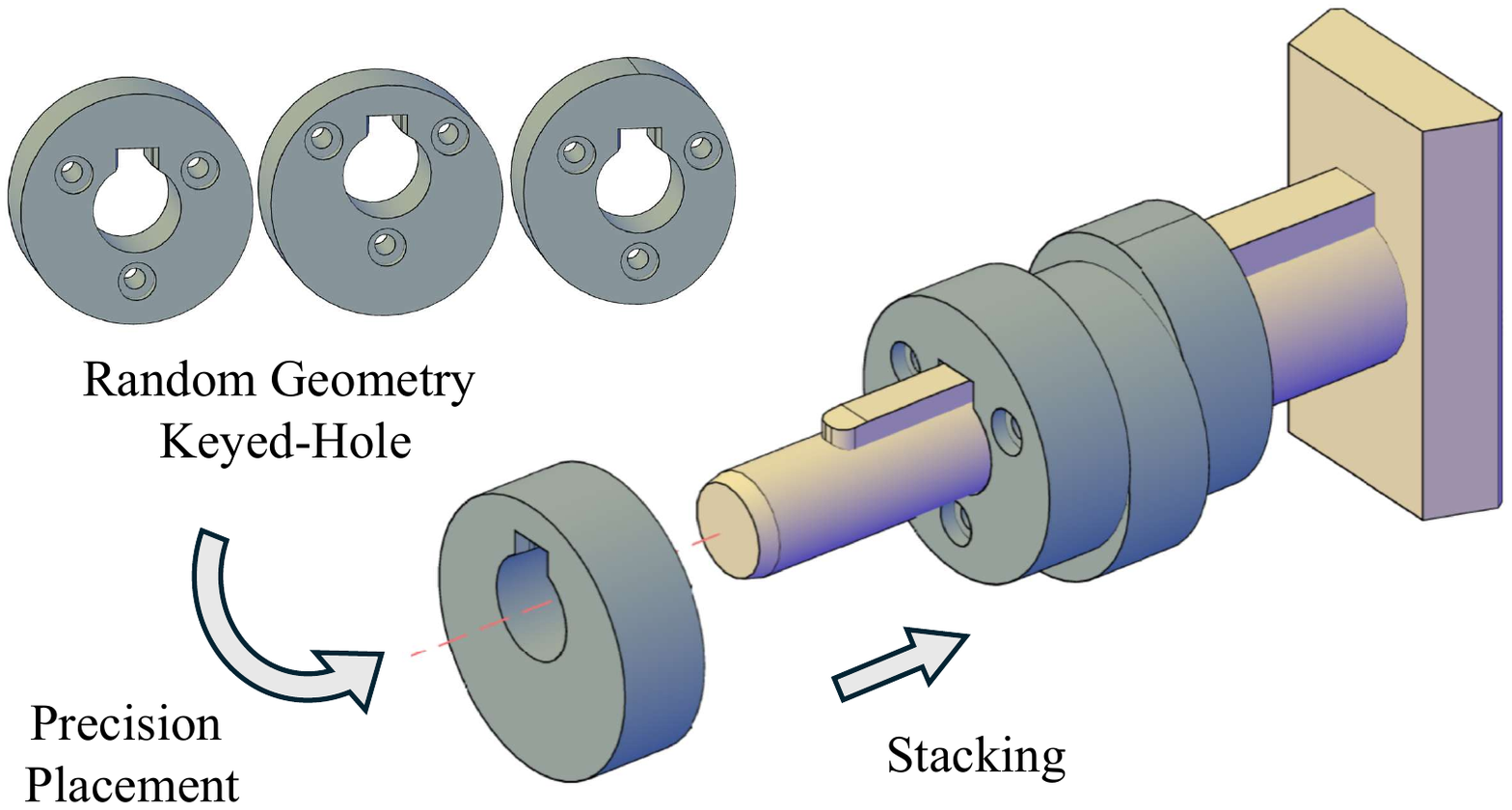}
    \caption{Insertion process with keyed circular hole.}
    \label{fig:keyed_circular_hole_insertion}
\end{figure}
\subsection{Related Work}
\vspace{-0.2em}
Force-based peg-in-hole alignment methods can be categorized into analytical contact models, contact-state estimation, or active search strategies driven by deterministic F/T feedback. 
Early analytical approaches, such as the closed-form three-point-contact alignment model, estimate insertion misalignment using known geometry and calibrated friction parameters~\cite{tang2016autonomous}, while more recent learning-based methods extend force-profile interpretation toward contact-manifold modelling and data-driven insertion correction~\cite{Sliwowski2025REASSEMBLE}.
Support Vector Machines (SVMs) and Gaussian Process (GP)-based approaches have demonstrated effective contact-state estimation and force-manifold learning. 
In the context of keyed circular assemblies, Huang et al.~\cite{huang2023hole} achieved 93--95\% deviation-state classification accuracy using GA-optimized SVMs trained across 180 samples per offset condition under discrete insertion states. 
In parallel, GP Regression (GPR) has been investigated for contact-force modelling and force-aware Model Predictive Control (MPC), to improve motion regulation under uncertainty~\cite{matschek2020direct}. 
Together, these works motivate local GPR-based continuous insertion-offset estimation from contact wrench signatures and pose-informed feedback.
\subsection{Problem Statement}
\begin{figure}[t]
    \centering
    \begin{subfigure}{\wfourdia\textwidth}
        \includegraphics[width=\linewidth]{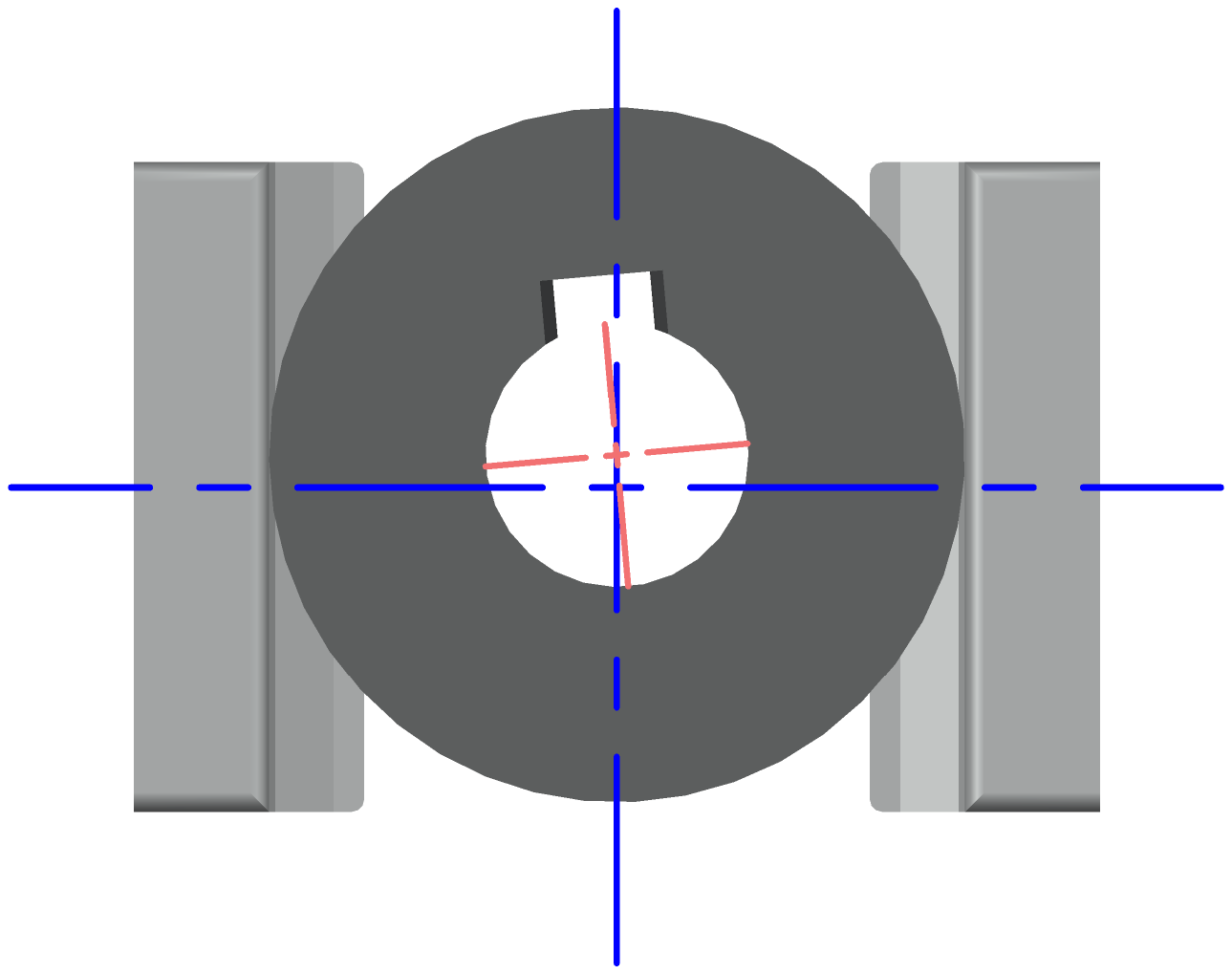}
        \vspace{-0.915cm}
        \subcaption{Rotational and radial \\ error, centered hole.}
        \label{fig:error_scenarios_during_insertion_a}
    \end{subfigure}
    \begin{subfigure}{\wfourdia\textwidth}
        \includegraphics[width=\linewidth]{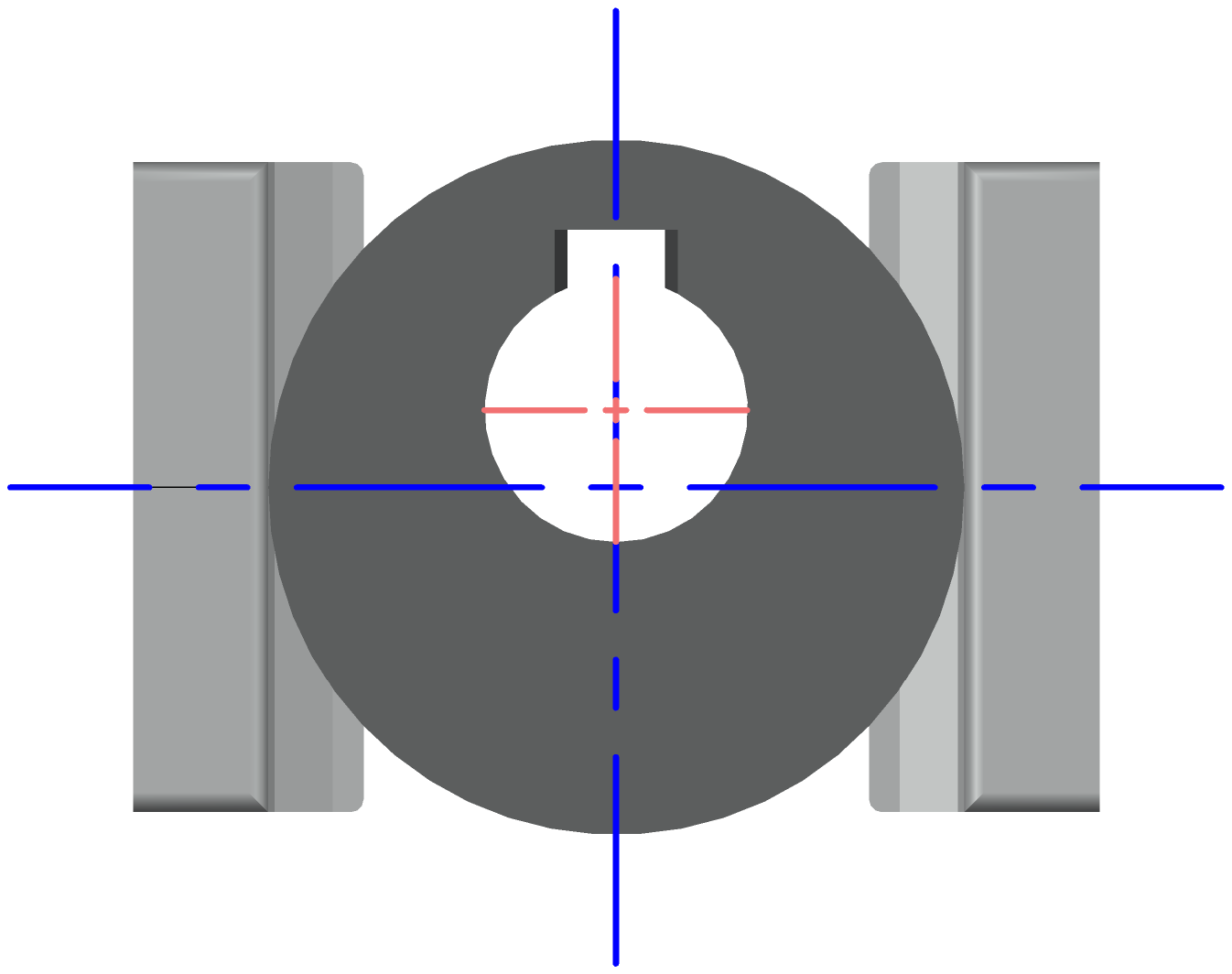}
        \vspace{-0.915cm}
        \subcaption{Radial error, off-centered hole.}
        \label{fig:error_scenarios_during_insertion_b}
    \end{subfigure}
    \begin{subfigure}{\wfourdia\textwidth}
        \includegraphics[width=\linewidth]{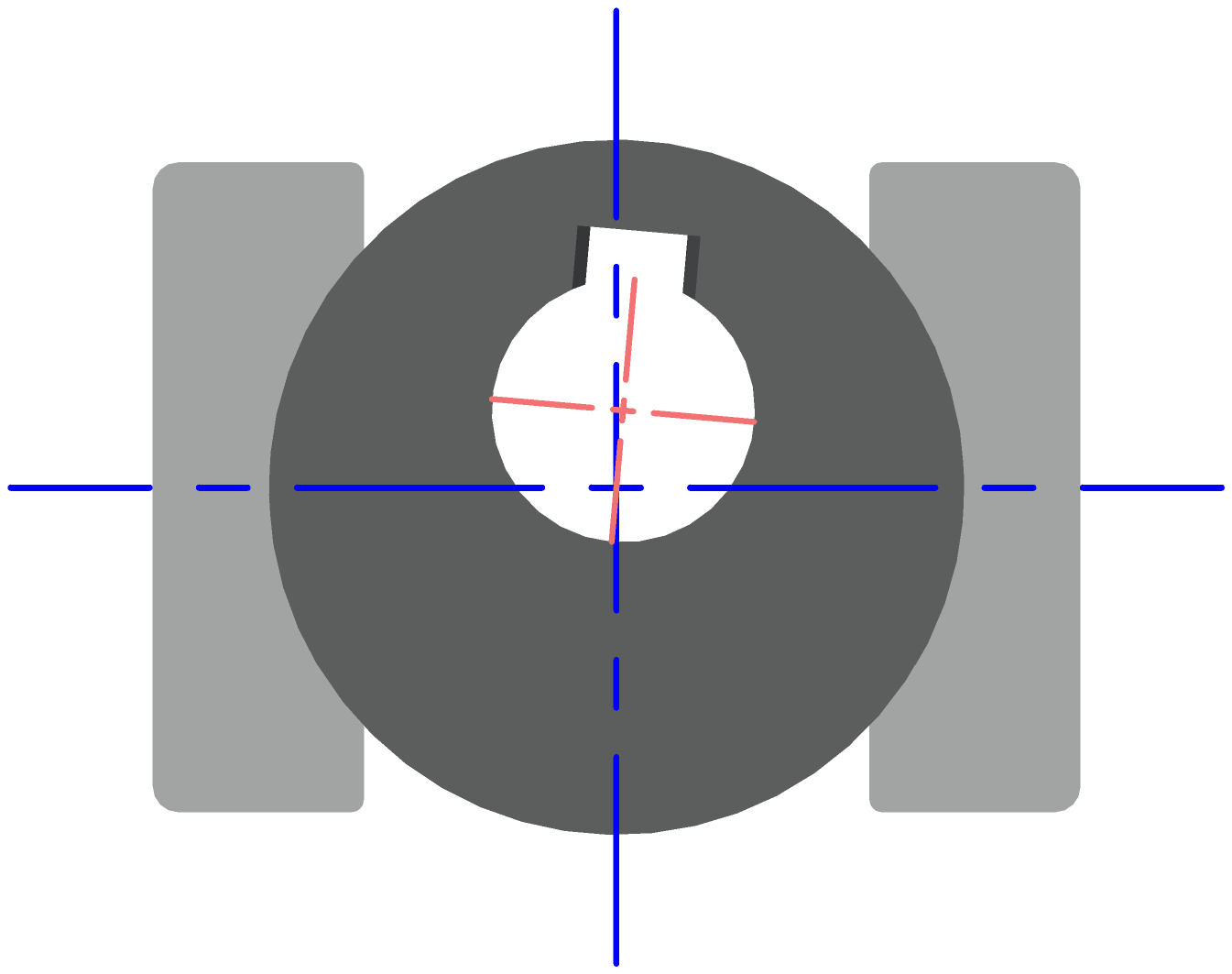}
        \vspace{-0.915cm}
        \subcaption{Rotational and radial error, off-centered hole,
                    mechanically fitted.}
        \label{fig:error_scenarios_during_insertion_c}
    \end{subfigure}
    \begin{subfigure}{\wfourdia\textwidth}
        \includegraphics[width=\linewidth]{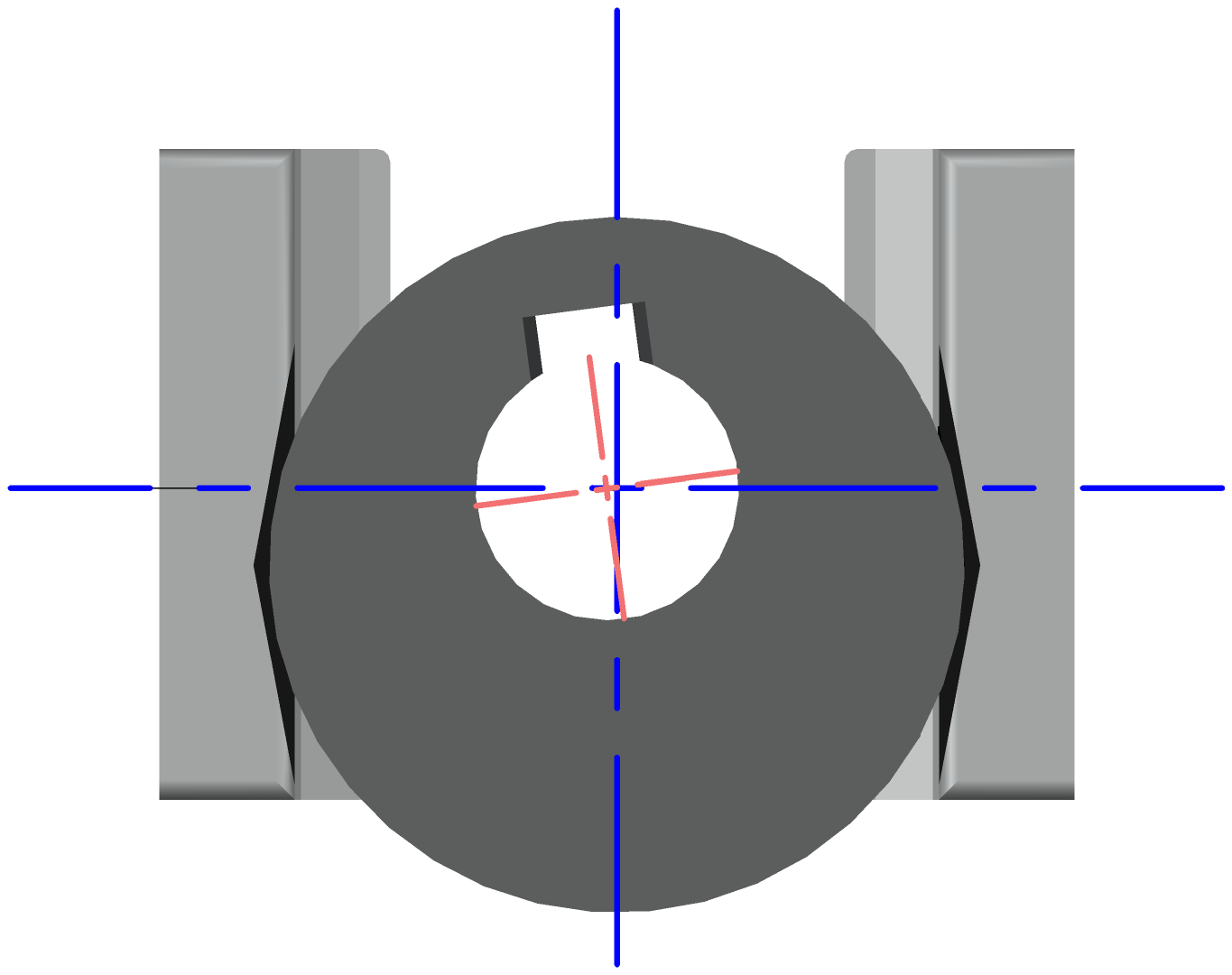}
        \vspace{-0.915cm}
        \subcaption{Rotational error, off-centered hole, mechanically fitted.}
        \label{fig:error_scenarios_during_insertion_d}
    \end{subfigure}
    \caption{Representative grasp-induced error configurations. Blue cross-hair: TCP; red cross-hair: keyed hole axis.}
    \label{fig:error_scenarios_during_insertion}
\end{figure}

The considered use-case is an industrial autonomous re-assembly process in which cylindrical discs with off-centered keyed holes and eccentric profiles are stacked onto a keyed shaft, as illustrated in Fig.~\ref{fig:keyed_circular_hole_insertion}, using a 6-axis collaborative robot arm.
The shaft includes a chamfer and a delayed key engagement, requiring correct alignment during initial insertion before final keyway coupling.
The peg--hole clearance is only $0.6\,\mathrm{mm}$.
In practice, grasping in cluttered environments introduces residual pose errors that propagate into the insertion stage.
Representative grasp-induced error configurations are shown in Fig.~\ref{fig:error_scenarios_during_insertion}.
Since the key engages only after full insertion, common search strategies such as spiral and keyway search, cannot be applied during initial contact.
Force-based reasoning therefore becomes the primary mechanism for detecting misalignment, but distinguishing hard collisions from chamfer-guided insertion is non-trivial.

\textbf{Research Questions.}
This work therefore addresses the following research questions for chamfered delayed key--keyway insertion: 
(\textbf{RQ1})~Can a regressor model estimate continuous planar insertion misalignment directly from wrist F/T measurements across hard-collision and chamfer-guided insertion contact regimes?
(\textbf{RQ2})~Can post-grasp visual pose validation improve force-guided insertion correction and trajectory re-parameterization under contact-rich assembly conditions?

\textbf{Contributions.} 
This work presents: (1) a local KNN--GP force-to-offset regression framework that estimates continuous residual planar misalignment for trajectory re-parameterization using visual post-grasp pose validation; and (2) a dual-regime insertion architecture distinguishing hard-collision and guided-insertion contact conditions during chamfer-guided keyed assembly.

\section{Pose-Validated Placement Pipeline}
\label{sec:systematic_pipeline}
\begin{figure}[t] 
    \centering
    \includegraphics[width=0.9\columnwidth]{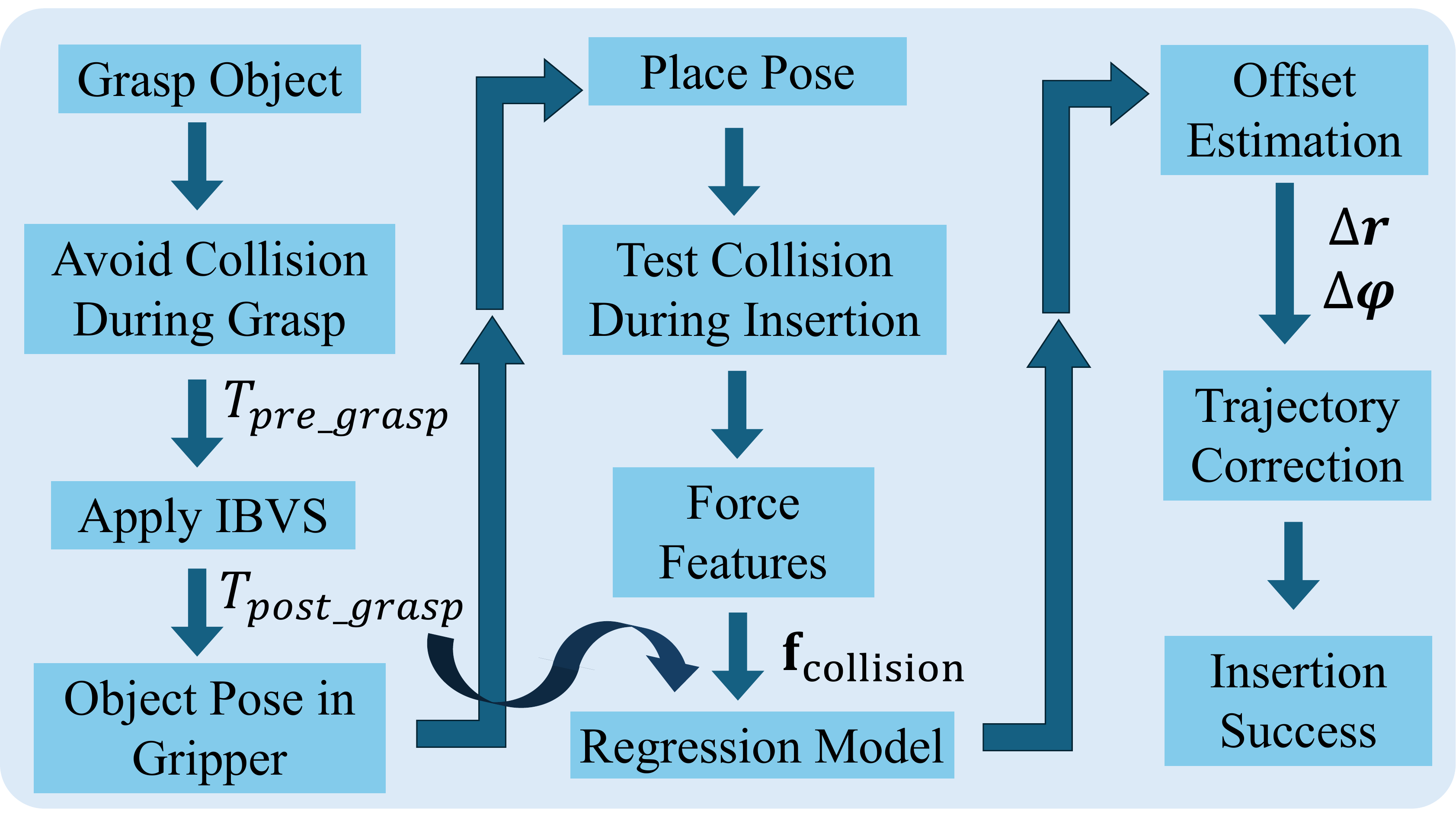}
    \caption{Complete pipeline for picking and insertion.}
    \label{fig:pipeline_diagram}
\end{figure}

The pipeline integrates monocular pose estimation, closed-loop planar pose regulation, and probabilistic force-based offset inference in three stages:
(1)~pre-grasp pose estimation,
(2)~post-grasp planar pose validation, and
(3)~F/T~-based trajectory correction.
The pipeline is illustrated in Fig.~\ref{fig:pipeline_diagram}.
\subsection{Pre-Grasp Pose Estimation}\label{subsec:pre_grasp_pose_estimation}
\begin{figure}[b] 
    \centering
    \hspace*{-1cm}
    \includegraphics[width=0.85\linewidth,trim=0 0 0 25,clip]{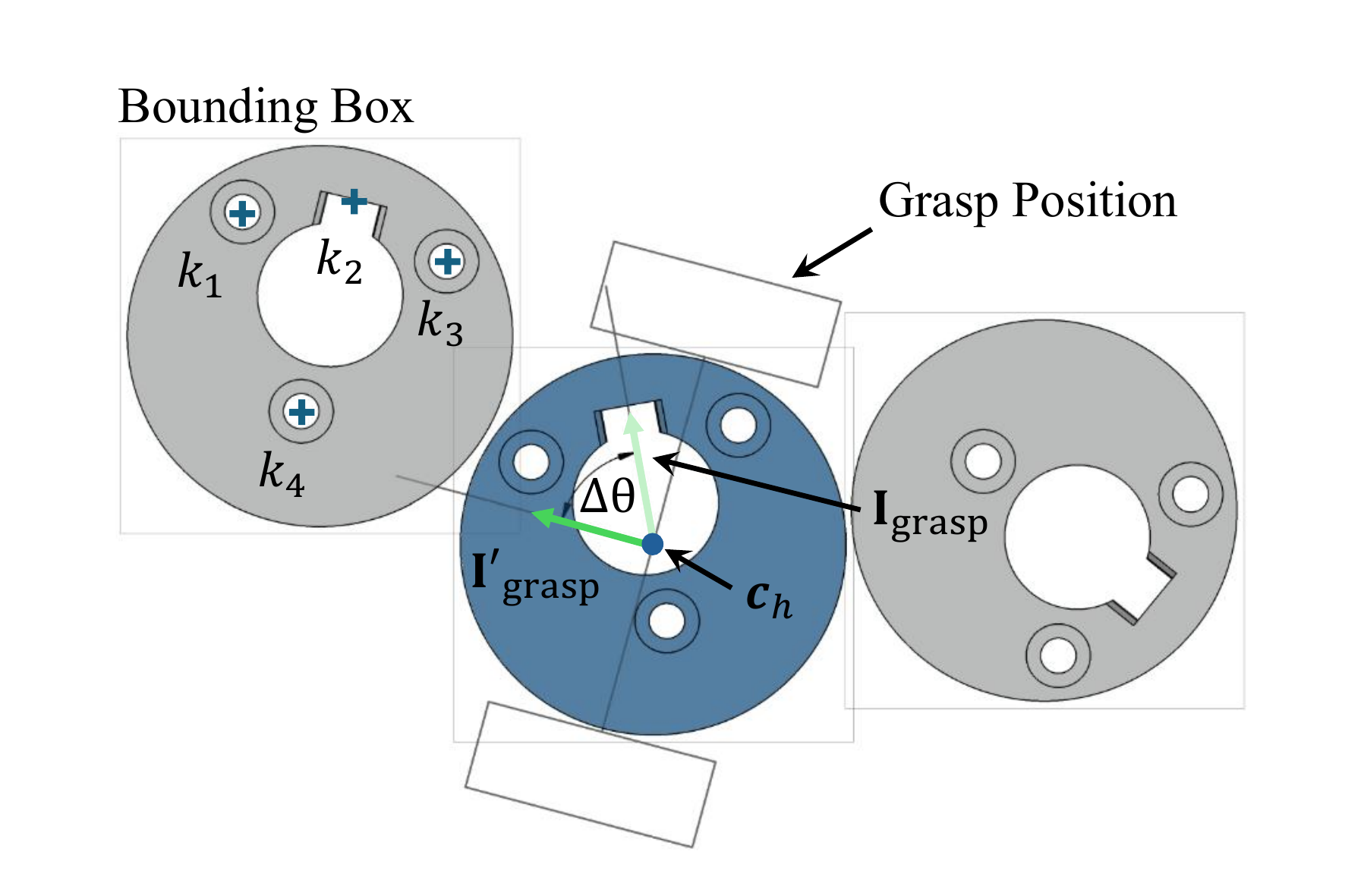}
    \caption{Collision-aware grasp planning. Blue disc: target; others: obstacles.}
    \label{fig:pre_grasp_collision_check_a}
\end{figure}

A custom-trained YOLOv8 pose-estimation model predicts bounding boxes and 2D keypoints ($k_1$--$k_4$): three structural reference points and one keyway feature (Fig.~\ref{fig:pre_grasp_collision_check_a}).
The hole center $\mathbf{c}_h$ is the mean of the structural
keypoints and the grasp orientation $\mathbf{l}_{\mathrm{grasp}}$ is the vector from the centroid to the keyway keypoint.
The gripper antipodal axis is checked for collision against neighbouring bounding boxes.
If detected, the grasp is iteratively rotated by $\Delta\theta$ until feasible, and the result is stored as $T_{\mathrm{pre\text{-}grasp}}$.

\subsection{Post-Grasp Planar Pose Validation}\label{subsec:post_grasp_planar_pose_val}
\begin{figure}[t] 
    \centering
    \includegraphics[width=1.0\columnwidth]{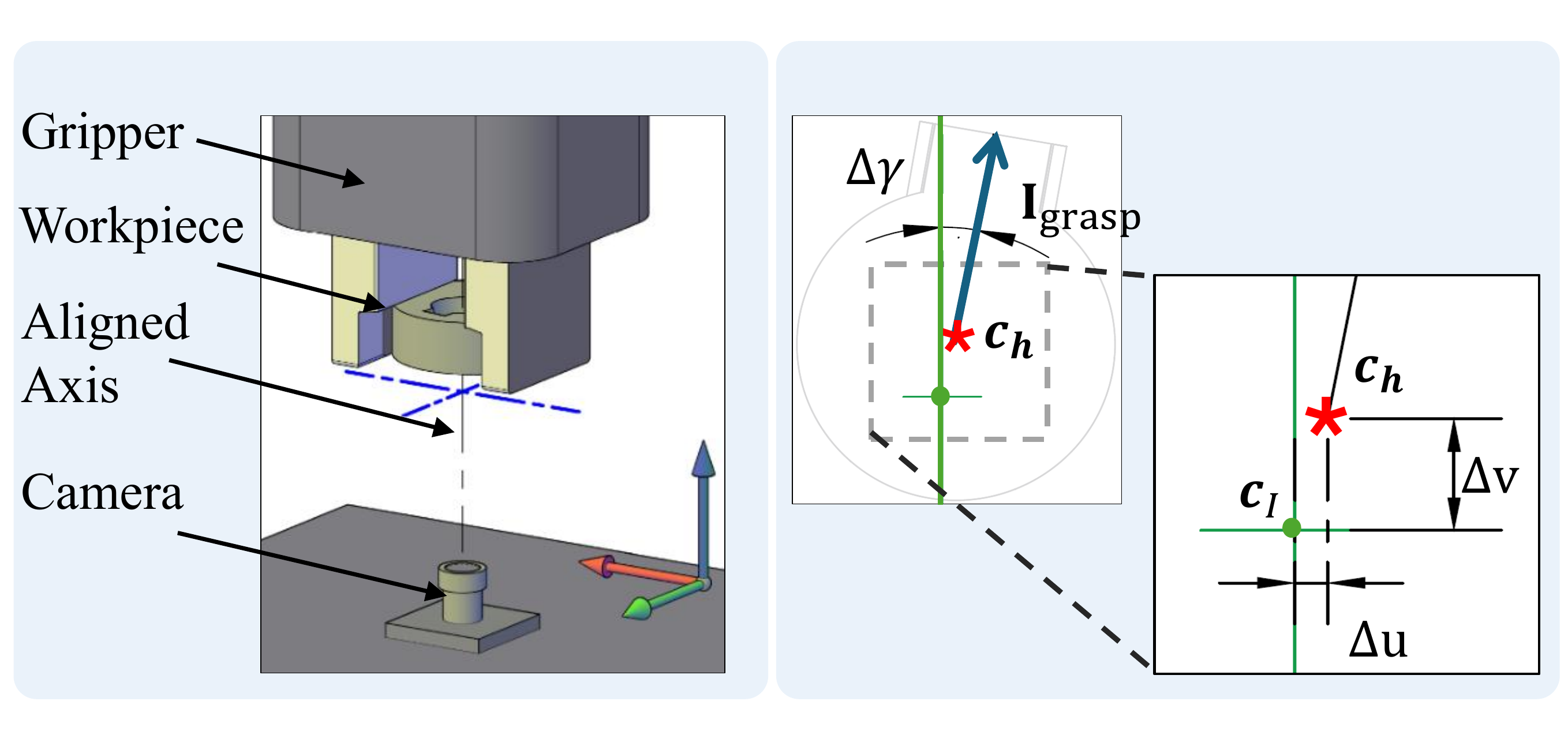}
    \caption{IBVS setup for post-grasp planar pose validation.}
    \label{fig:check_pose_ibvs_figure}
\end{figure}

Rather than re-estimating the full $SE(3)$ pose after grasping, a task-constrained planar regulation approach,  $SE(2)$, based on Image-Based Visual Servoing~(IBVS) is adopted.
The manipulator moves to a predefined waypoint above an externally mounted, calibrated camera whose optical axis is aligned with the TCP, as shown in Fig.~\ref{fig:check_pose_ibvs_figure}.
The transform $T_{\mathrm{pre\text{-}grasp}}$ is applied to reduce the initial pixel misalignment $(\Delta u, \Delta v)$ between the detected hole center and the image center, $c_I$, as well as the angular offset $\Delta \gamma$ between $\mathbf{l_{grasp}}$ and the image vertical axis.
Convergence, and thus IBVS success, is declared when:
\begin{equation}
    |\Delta u| \leq 4\,\text{pixels}, \quad |\Delta v| \leq 4\,\text{pixels}, \quad |\Delta\gamma| \leq 0.5^\circ
\label{eq:ibvs_convergence}
\end{equation}

The misalignments are regulated to zero via closed-loop IBVS.
With $T_0$ denoting the TCP pose at the onset of IBVS and $T_f$ the converged pose, the validated planar correction (relative transform) is given by
\begin{equation}
T_{post-grasp} = T_0^{-1} T_f =
\begin{bmatrix}
R & \mathbf{p} \\ 0 & 1
\end{bmatrix}.
\label{eq:check_pose_transform}
\end{equation}
Because $T_{post-grasp}$ is derived from encoder-measured displacement after visual convergence, it constitutes a metric correction exploiting the sub-millimeter repeatability of modern manipulators.

\begin{figure}[b] 
    \centering
    \includegraphics[width=0.85\columnwidth]%
      {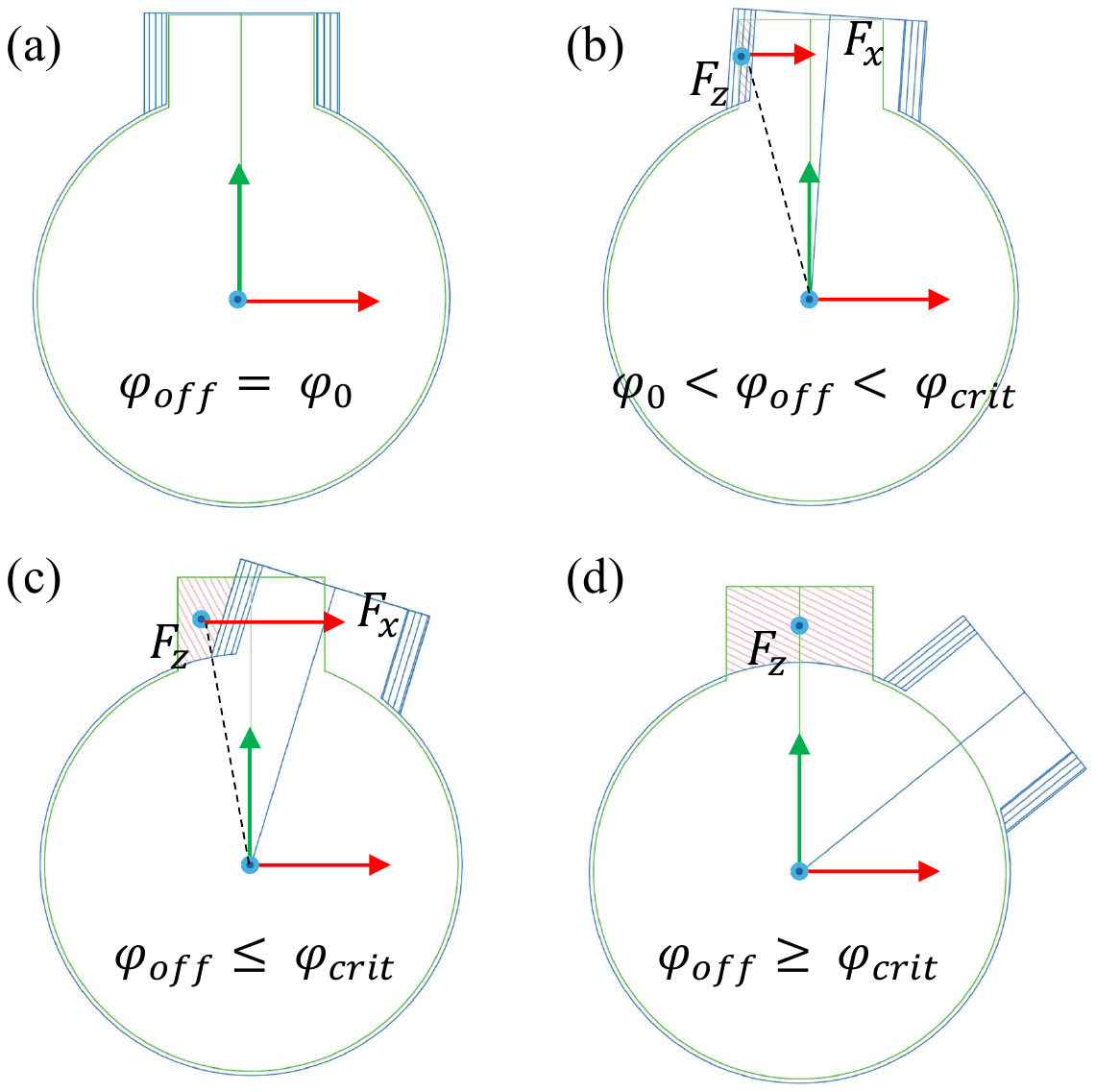}
    \caption{Key--keyway contact wrench distribution during disc insertion under angular misalignment.}
    \label{fig:insertion_error_front_view}
\end{figure}
\subsection{Probabilistic Force/Torque-Based Trajectory Correction}\label{subsec:probabilistic_force_torque_trajectory}
\begin{figure}[b]
    \centering
    \includegraphics[width=0.85\columnwidth]%
      {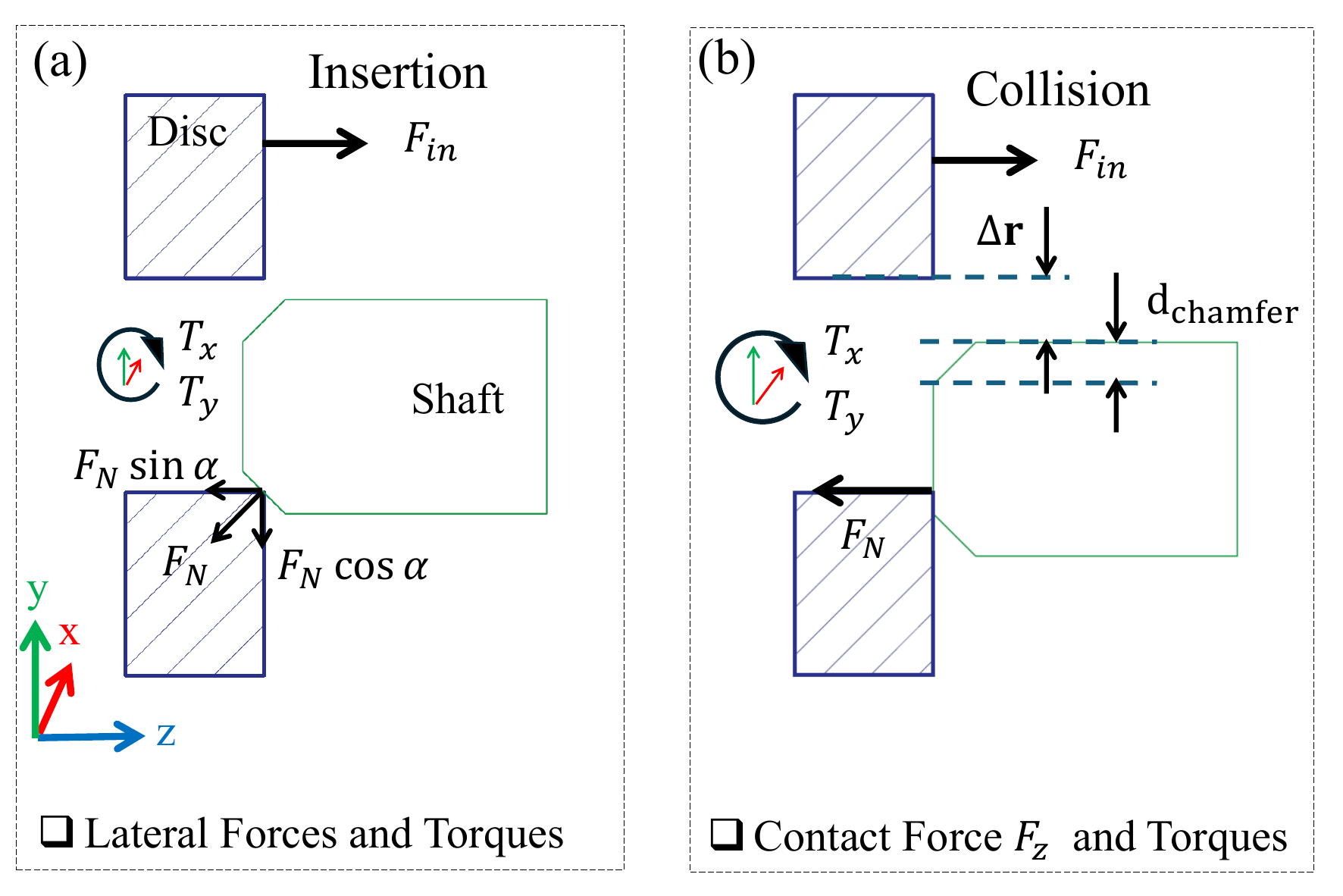}
    \caption{Contact wrench distribution under radial misalignment during insertion (z-axis).}
    \label{fig:insertion_error_radial}
\end{figure}

The disc is then placed at the end of the keyed shaft.
After applying the correction $T_{\text{post-grasp}}$ to align the hole center with the keyed shaft, insertion is initiated.
The regression target is the residual planar offset $\boldsymbol{\Delta r} = [\Delta x,\, \Delta y]^T$, as well as the angular offset $\mathbf{\varphi}_{\mathrm{off}}$, quantifying misalignment in the insertion plane, as illustrated in Fig.~\ref{fig:insertion_error_front_view} and Fig.~\ref{fig:insertion_error_radial}.
Residual misalignment generates characteristic contact wrenches $\mathbf{w} = [F_x, F_y, F_z, \tau_x, \tau_y, \tau_z]^T$ at the robot wrist.

\subsubsection*{Contact Regime Classification}
Two distinct contact scenarios arise depending on offset magnitude relative to the shaft chamfer.
(a)~\textit{Guided insertion}: When the offset is small enough for the chamfer to capture the disc, a sustained force profile is generated.
(b)~\textit{Hard collision}: When the offset exceeds that radial length of the chamfer, $\|\boldsymbol{\Delta r}\| > d_{\mathrm{chamfer}}$, the disc impacts the shaft rim and the robot halts due to the force limit along its $TCP~z-axis$, producing a short transient contact.
For the key--keyway configuration, guided insertion is feasible for $\mathbf{\varphi}_{\mathrm{off}} < \pm 6^\circ$. 
At larger offsets ($\pm 8^\circ$--$\pm 11^\circ$), partial collisions occur before reaching the  critical angle $\mathbf{\varphi}_{\mathrm{crit}}$, resulting in ~\textit{hard collision}.
These regimes yield fundamentally two different sets of force signatures, motivating a dual-model architecture.

\begin{figure}[b]
    \centering
    \includegraphics[width=0.85\columnwidth]%
    {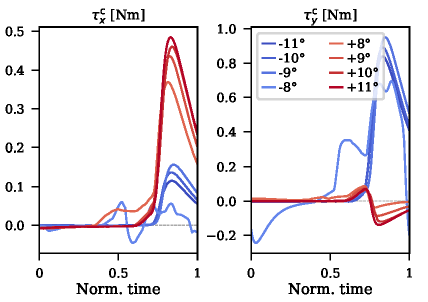}
    \caption{Contact torque signature comparison over a number of trials for insertion scenarios.}
    \label{fig:contact_regime_comparison}
\end{figure}

During contact with a chamfer inclined at an angle $\alpha$, radial misalignment induces a contact force with axial component $F_y \propto F_N\cos\alpha$ and lateral component $F_z \propto F_N\sin\alpha$. 
During guided insertion, angular misalignment between the key and keyway produces asymmetric torque signatures about $\tau_z$, while residual radial offset contributes to $\tau_x$ and $\tau_y$. 
Under larger angular deviations, the contact transitions toward unstable edge interaction and collision-dominant behaviour, resulting in less consistent rotational torque signatures. 
In hard-collision conditions, distinct wrist torque vectors arise from eccentric contact forces acting through the offset-dependent moment arm between the contact point and the wrist frame.
Fig.~\ref{fig:contact_regime_comparison} demonstrates contact torque signature during insertion.
\subsubsection*{Force Feature Representation}
For each $F/T$ channel $c$ over the detected contact window $N_{contact}$, five summary statistics are extracted: mean, Root Mean Square (RMS), maximum, minimum, and standard deviation.
For guided insertion, all six channels yield a 30-dimensional feature vector~$\mathbf{f}_{\mathrm{ins}}$; for hard collision, only $\tau_x$ and $\tau_y$ are used, yielding a 10-dimensional vector~$\mathbf{f}_{\mathrm{col}}$. 
For angular offset, $\mathbf{f}_{\mathrm{col}}$ is used, until $\mathbf{\varphi}_{\mathrm{off}}$ approaches $\mathbf{\varphi}_{\mathrm{crit}}$, after which, the key--key-way collision triggers no signficant force signatures, and post-grasp validation has to be reiterated.
The torque channels directly encode the contact moment arm.
The feature vector~$\mathbf{f}_{\mathrm{ins}}$ validates successful insertion after correction and confirms offset adequacy, since the chamfer continues to exert radial wrist forces during guided entry.
\subsubsection*{Local KNN--GP Hybrid Regression}
\begin{figure}[b]
    \centering
    \includegraphics[width=1.0\columnwidth]%
      {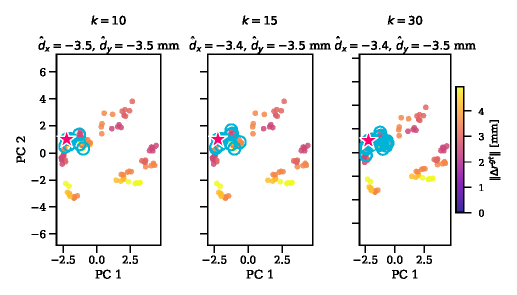}
    \caption{PCA projection of force-feature space, coloured by $\|\boldsymbol{\Delta r}^{gt}\|$. Star: query; circles: $k$ nearest neighbours.}
    \label{fig:force_manifold_knn}
\end{figure}

A GP model approximates the nonlinear mapping
\begin{equation} \label{eq:gp_full}
\boldsymbol{\Delta r} = g(\mathbf{f}) + \boldsymbol{\epsilon}, 
\qquad
g(\mathbf{f}) \sim \mathcal{GP}\!\left(\mu(\mathbf{f}),\,k(\mathbf{f},\mathbf{f}')\right),
\end{equation}
where $g(\mathbf{f})$ is modelled as a GP with mean function $\mu(\mathbf{f})$ and covariance kernel $k(\mathbf{f},\mathbf{f}')$.
The mapping is modeled using a Squared Exponential kernel with Automatic Relevance Determination (ARD) \cite{williams2006gaussian}. 
While the ARD kernel improves flexibility, a \emph{global} GP exhibits smoothing-induced bias in sparse regions.
Therefore, for each query $\mathbf{f}_*$, the $k=30$ nearest neighbours are retrieved in normalized force-space, and a \emph{local} GP is fitted on this subset only, separately for the planar components of $\boldsymbol{\Delta r}$, restricting inference to the relevant region of the force manifold.
To ensure local consistency, the IBVS-based offset estimate $T_{\mathrm{post-grasp}}$
is incorporated as an informative prior in the GP model, which predicts a residual correction relative to this estimate, while neighbor retrieval remains based solely on the force-feature space.
Its translational component provides an offset estimate ${T}_{\mathrm{post-grasp}} = \mathbf{p}_{xy}$ from Eq.~\eqref{eq:check_pose_transform}.
\noindent The PCA projection in Fig.~\ref{fig:force_manifold_knn} illustrates the local structure of the force–feature manifold, where neighbouring samples exhibit similar ground-truth radial offsets, $\boldsymbol{\Delta r}^{gt}$. 
Predictions remain stable for neighbourhood sizes $k \in \{10,15,30\}$, supporting the assumption of a locally smooth mapping between contact wrench signatures and planar misalignment.

\subsubsection*{Trajectory re-parameterization}
The final predicted offset $\hat{\boldsymbol{\Delta r}}$ is then applied as a supplemental translational correction:
\begin{equation}
T_{\text{pred}} = 
\left( T_{\text{shaft}} T_{\text{post-grasp}} \right)
\begin{bmatrix}
I_{3\times3} & \hat{\boldsymbol{\Delta r}} \\ 0 & 1
\end{bmatrix},
\label{eq:trajectory_update}
\end{equation}
yielding a deterministic trajectory re-parameterization at the moment of contact, at the shaft pose $T_{shaft}$.
\section{Experimental Setup}\label{sec:experimental_setup}

\begin{figure}[t] 
    \centering
    \includegraphics[width=1.\columnwidth]{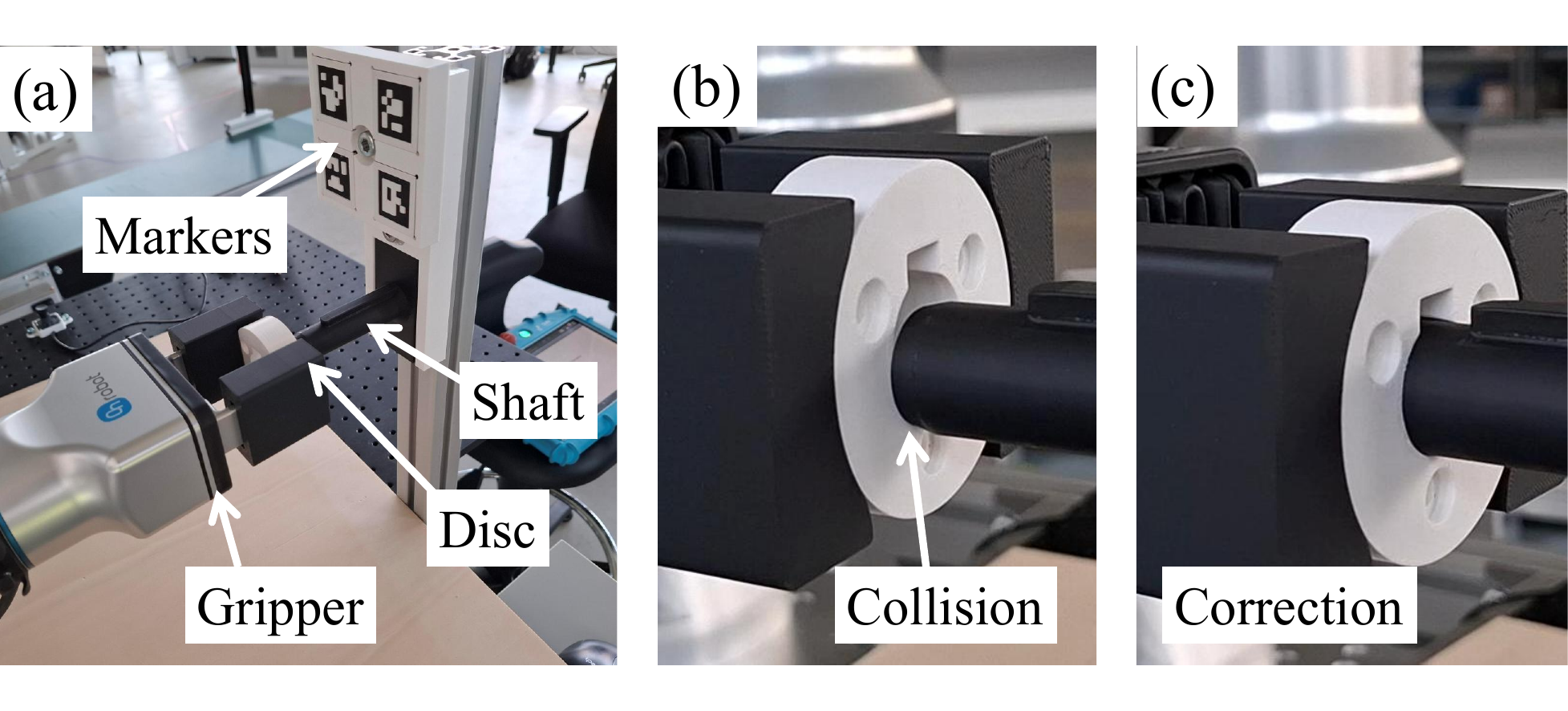}
    \caption{(a) Gripper and disc setup, with ArUco markers for shaft localization, (b) collision during placement, (c) trajectory correction applied.}
    \label{fig:hardware_setup}
\end{figure}

The setup replicated an industrial key--keyway mounting scenario, executed as \textit{grasp} $\rightarrow$ \textit{check} $\rightarrow$ \textit{place} with parallel $F/T$ logging.
Ground truth relied on the planar offset $\boldsymbol{\Delta r}^{gt}$ and angular offset $\varphi^{gt}$.
The primary metrics were the predicted offsets and the insertion success rate.
Jamming during precise-placement constituted a failed insertion.
The insertion success criteria was stated as $\|\boldsymbol{\Delta r}\| \leq d_{\mathrm{chamfer}}$.
As shown in Fig.~\ref{fig:hardware_setup}(a), a \textit{Universal Robots} UR5e manipulator with integrated wrist $F/T$ sensing was fitted with mechanical centering gripper fingers. 
The UR5e has a wrench sensor with $\pm3.5~N$ force and $\pm0.2~Nm$ torque accuracy.
The keyed shaft was localized via four coplanar ArUco markers.
A wrist-mounted RealSense D415 provided RGB imagery for pose estimation and depth for workpiece and shaft localization.
The shaft chamfer had an inclination $\alpha ~\text{of}~15\degree$ with axial length of $4\,\mathrm{mm}$ thus $d_{\mathrm{chamfer}}\,=\,1.07~mm$.
A custom-trained YOLOv8 pose-estimation model, trained with $30\,000$ synthetically generated images, was used for keypoint-based grasp initialization.
During IBVS, keypoint jitter was suppressed by a Kalman filter ($Q = 10^{-9}$, $R = 0.3$) at $25\,\text{Hz}$, prioritizing physical motion consistency over detector noise.
Insertion proceeded at a creep speed of $0.01\,\mathrm{m/s}$ after taring the sensor and recording a $2.0\,\mathrm{s}$ static baseline to eliminate $F/T$ bias and acceleration artifacts.
Contact onset was detected by simultaneous $F/T$ value increase and velocity reduction which were registered as data rows through the integrated Real-Time Data Exchange (RTDE) protocol of Universal Robots at $125\,\mathrm{Hz}$.
A force limit along the insertion axis, $|F_z| > F_{\text{th}}$ with $F_{\text{th}} = 30\,\mathrm{N}$ (selected over 5~N and 15~N thresholds for cleaner regime separation), routed hard collisions ($<$100 post-onset data rows) to the GP-collision model, while completed insertions ($\geq$100 data rows) were passed to the GP-insertion model.
The KNN--GP was implemented in \texttt{scikit-learn} with an ARD--RBF kernel.

For each workpiece a dataset of 325 data samples were collected, containing the corresponding feature vectors~$\mathbf{f}_{\mathrm{ins}}$ and $\mathbf{f}_{\mathrm{col}}$, as well as $\boldsymbol{\Delta r}^{gt}$ and $\varphi^{gt}$.
An automated sweep controller iterated the robot waypoints over a predefined grid of planar offsets, executing five repeat trials per condition and logging synchronized $F/T$ values, TCP pose and velocity via RTDE.
Table~\ref{tab:dataset_summary} summarizes dataset configurations, where each conditions had 5 trials.
Algorithm~\ref{alg:trajectory_reparametization} shows the implemented trajectory re-parameterization procedure.

\begin{table}[t] 
\centering
\small
\caption{Dataset Summary}
\label{tab:dataset_summary}
\renewcommand{\arraystretch}{1.05}
\begin{tabular}{lccc}
\toprule
 & Insertion & Collision & Angular \\
\midrule
Trials & 125 & 160 & 40 \\
Conditions & 24 & 32 & 8 \\
Radial offset (mm) & 0.5–1.5 & 2.0–3.5 & -- \\
Angular offset ($^\circ$) & -- & -- & $\pm{8–11}$ \\
Feature dim. & 30 (6 ch.) & 10 (2 ch.) & 10 (2 ch.) \\
Contact data rows & $\geq100$ & $<100$ & $<100$ \\
\bottomrule
\end{tabular}
\end{table}
\section{Experimental Results and Discussion}
\label{sec:experimental_results_discussions}

The validation experiments were conducted over 30 trials of pick, post-grasp validation, and placement using three workpiece variants with hole offsets between $0$ and $8\,\mathrm{mm}$ relative to the workpiece centre. 
To ensure a controlled comparison between conditions, only the correction strategy (Baseline, GP, GP+KNN, and GP+KNN+Mean Prior) was varied between runs, while the shaft localisation data and task execution order remained fixed.
The baseline relied solely on pose estimation and ArUco-based shaft localisation without IBVS or force-based correction.
Across all conducted experiments, the pose-estimation model achieved approximately $92\%$ keypoint detection success, while post-grasp IBVS converged in approximately $95\%$ of trials according to Eq.~\eqref{eq:ibvs_convergence}. 
Differences in insertion performance therefore primarily reflect the contribution of the correction framework rather than variability in perception or manipulation.
\subsection{Baseline Condition without Offset Regression}\label{subsec:baseline_condition_without_offset}
Of the 30 baseline trials, 20 resulted in successful insertion, corresponding to a baseline insertion success rate of $67\%$.
Most failures originated from hard collisions between the shaft and disc caused by residual hand--eye calibration error and camera-to-TCP transformation uncertainty. 
Inaccurate keypoint detection resulted in invalid pose estimates in 2 trials, while pose-estimation and calibration errors introduced additional insertion misalignment in 8 trials, leading to inaccurate localization of the hole pose within the gripper frame.

\begin{algorithm}[t]
\caption{Force-based Contact Correction for Insertion}
\label{alg:trajectory_reparametization}
\begin{algorithmic}[1]
\State Move robot to pre-insertion pose $T_{shaft}$
\State Start force/torque data collection
\Repeat
    \State Execute nominal insertion along TCP axis $\mathbf{z}$
    \State Monitor rolling feature window $\mathbf{f}(t)$
    \If{$F_z > F_{\text{th}}$ (contact detected)}
        \State Stop motion at contact pose $\mathbf{p}$
        \State Extract $N_{contact}$ from recorded data
        \State Calculate $T_{\text{pred}}$ using Eq.~\ref{eq:gp_full} and Eq.~\ref{eq:trajectory_update}
        \State Retract along insertion axis
        \State Update insertion trajectory:
        \[
        \mathbf{p}'(s)=\mathbf{p}+T_{\text{pred}}+s\,\mathbf{z}
        \]
        \State Resume along re-parameterized trajectory
    \Else
        \State Continue insertion
        \State Monitor rolling feature window for insertion
    \EndIf
\Until{successful insertion or maximum retries reached}
\State Finish insertion
\end{algorithmic}
\end{algorithm}
\vspace{-0.55em}
\subsection{Force-Based Offset Regression}\label{subsec:force_based_offset_regression}
The full pipeline, with and without post-grasp IBVS validation and GPR-based trajectory correction (GP$+$KNN$+$Mean Prior), was evaluated over the same 30 trials under identical conditions. 
A mean of 26 of the 30 trials resulted in successful insertion ($87\%$), with trajectory re-parameterisation completed within two correction attempts in all successful cases. 
Failures primarily occurred when insertion offsets exceeded the trained manifold boundary of $\pm3.5\,\text{mm}$ after repeated collision attempts, where the GP could no longer reliably extrapolate from the available force-feature gradients. 

\begin{figure}[b]
    \centering
    \hspace*{-1cm}
    \includegraphics[width=0.75\columnwidth]{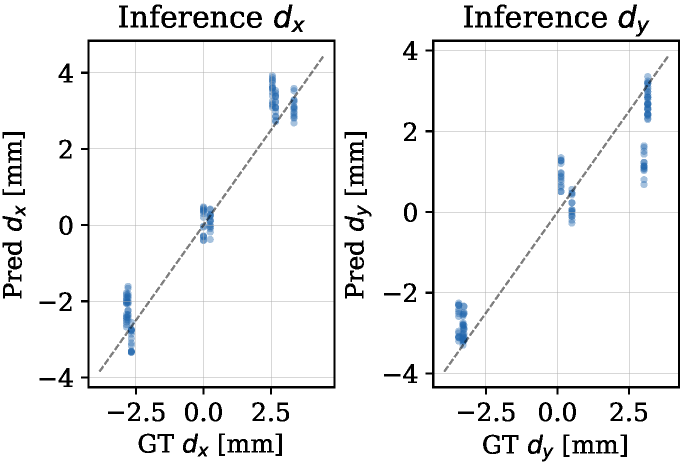}
    \caption{Predicted vs.\ ground-truth planar offsets and cumulative error distribution.}
    \label{fig:predicted_vs_gt}
\end{figure}

These failures originated mainly from pose-estimation inaccuracies propagating beyond the trained force-manifold range.
Fig.~\ref{fig:predicted_vs_gt} shows the predicted versus ground-truth offsets estimated by the GP--KNN model.
Predictions cluster around discrete offset conditions within $[-3.5,\,3.5]\,\text{mm}$, confirming accurate planar misalignment recovery in cases where the initial pose estimation was insufficient for direct insertion. 
Extending the training dataset range to $\pm4.5\,\text{mm}$ resolved these failure cases; however, these trials were retained in the reported success-rate evaluation over the full 30 trials. 
As the insertion offset approaches the boundary regions of the chamfer capture range, the interaction transitions from sustained chamfer-guided contact toward transient edge-impact behaviour, producing shorter and less distinctive contact-wrench signatures. 
This reduces the local feature separability within the force manifold and limits reliable GP inference near the extrapolation boundary.

\subsubsection{Gaussian Process Regression}
The performance of the dual-model architecture is summarized in Table~\ref{tab:loco_results}, demonstrating the effectiveness of the local KNN--GP regressor in resolving residual planar misalignments.
The mean inference latency was $472\,\text{ms}$, requiring a minimum contact window of 80 data rows ($640\,\text{ms}$ at $125\,\text{Hz}$) before inference.
The Leave-one-angle-out cross-validation (LOCO) evaluation assessed generalization to held-out offset conditions within the dataset, showing that the proposed KNN-GP with mean prior achieved the lowest average and tail errors across all metrics. 
Although the evaluated workpieces included hole offsets up to $8.0\,\mathrm{mm}$, the IBVS-derived GP prior reduced prediction errors by guiding inference toward geometrically consistent corrections within the typical insertion range.
When the predicted correction exceeded the chamfer capture threshold, $|\mathbf{e}_{\text{pred}}| > d_{\mathrm{chamfer}}$, indicating unreliable estimates, the system fell back to the nearest-neighbour offset to ensure a stable correction, together with the mean prior.

\begin{table}[t]
\centering
\caption{LOCO cross-validation results evaluating generalization to held-out offset conditions.}
\label{tab:loco_results}
\renewcommand{\arraystretch}{1.15}
\begin{tabular}{@{}lccc@{}}
\toprule
Metric & GP & GP+KNN & GP+KNN+Mean Prior \\
\midrule
MAE $d_x$ & $0.535$~mm & $0.520$~mm & $\mathbf{0.458}$~mm \\
MAE $d_z$ & $1.004$~mm & $\mathbf{0.721}$~mm & $0.734$~mm \\
MAE $|\mathbf{e}|$ & $1.216$~mm & $0.972$~mm & $\mathbf{0.947}$~mm \\
RMSE $|\mathbf{e}|$ & $1.487$~mm & $1.182$~mm & $\mathbf{1.148}$~mm \\
Median error & $0.972$~mm & $0.729$~mm & $\mathbf{0.719}$~mm \\
90th-percentile error & $2.441$~mm & $1.889$~mm & $\mathbf{1.804}$~mm \\
\bottomrule
\end{tabular}
\end{table}
\subsubsection{Angular Offset Estimation}
Leave-one-angle-out cross-validation yielded an overall MAE of $0.97^{\circ}$ and RMSE of $1.04^{\circ}$, with per-condition MAE ranging from $0.61^{\circ}$ ($\varphi^{gt} = {+}9^{\circ}$) to $1.32^{\circ}$ ($\varphi^{gt} = {-}8^{\circ}$).
For $\varphi$ values between $\pm1~^\circ$ and $\pm7~^\circ$, since the fixture incorporates a chamfered entry and an automatic wrist-roll correction stage, residual angular errors were mechanically absorbed during final seating, and no placement failures were observed across all evaluated trials.

\subsection{End-to-End Insertion and Discussion}\label{subsec:end_to_end_insertion}
Two principal findings emerge: 
(1)~The GP regression reliably provides continuous radial and angular offset estimates using a total of fewer than 400 training samples, unlike the GA-SVM which requires 180 samples per deviation and is limited to discrete classification.
(2)~The local KNN--GP with ARD kernel achieves highest accuracy without CAD-based contact modeling, by constraining inference to locally similar force signatures and fusing with the IBVS prior.
The mean prior improved prediction accuracy and robustness near the manifold boundaries.

Furthermore, the dual-model architecture extends estimation to hard-collision scenarios via contact-duration routing and torque-only features, recovering accurate estimates from transient ($\approx 1.0\,\mathrm{s}$) events with fewer attempts, critical for preventing repeated insertion attempts that could damage hardware.
Insertion success increased from 67\% to 87\%.
It should also be noted that while a YOLOv8 pose estimator and IBVS-based post-grasp validation are employed in this work, the GP regression framework depends only on the metric correction $T_{\text{post-grasp}}$ as an informative mean prior and is agnostic to the specific visual pipeline used to derive it.

One major limitation is the out-of-distribution scenario, as the GP cannot extrapolate beyond the training envelope.
The integrated wrist \mbox{F/T} sensor of the UR5e ($\pm3.5\,\text{N}$ force accuracy, $\pm0.2\,\text{Nm}$ torque accuracy) imposes a noise floor that bounds the minimum resolvable offset: misalignments generating contact wrenches below this threshold cannot be reliably distinguished from sensor noise.
Dedicated external \mbox{F/T} sensors would improve overall metric measurement accuracy.

\section{Conclusion}\label{sec:conclusion}

This work presented a structured precision placement framework for delayed key--keyway assemblies, integrating pre-grasp keypoint-based pose estimation, post-grasp IBVS validation, and force-driven trajectory re-parameterization.
The proposed local KNN--GP regressor estimates planar misalignment directly from contact force features, replacing discrete classification with continuous force-to-offset mapping.
Experimental results show that post-grasp validation improves insertion reliability from 67\% to 87\%, while the dual-model architecture achieves mean errors below $1\,\mathrm{mm}$ from both guided-insertion and transient collision events across continuous offset conditions.
The probabilistic offset estimate provides a principled conditioning variable for future integration with adaptive movement primitives.

\bibliographystyle{IEEEtran}
\bibliography{references}

\end{document}